\documentclass[a4paper]{article}

\usepackage{INTERSPEECH2015}

\usepackage{graphicx}
\usepackage{amssymb,amsmath,bm}
\usepackage{textcomp}

\ninept

\title{Maximum a Posteriori Adaptation of Network Parameters in Deep Models}

\makeatletter
\def\name#1{\gdef\@name{#1\\}}
\makeatother
\name{{\em Zhen Huang$^1$, Sabato Marco Siniscalchi$^{1,2}$, I-Fan Chen$^1$, Jinyu Li$^3$, Jiadong Wu$^1$, and Chin-Hui Lee$^1$}}
\address{$^1$School of ECE, Georgia Institute of Technology, Atlanta, GA. USA \\
  $^2$Department of Computer Engineering, Kore University of Enna, Enna, Italy \\
  $^3$ Microsoft Corporation, One Microsoft Way, Redmond, WA. USA \\
  {\small \tt huangzhenee@gatech.edu, marco.siniscalchi@unikore.it, ichen8@gatech.edu,}\\
   {\small \tt jinyli@exchange.microsoft.com, jwu65@gatech.edu, chl@ece.gatech.edu}
}

\begin{document}
%\ninept
%
\maketitle
\begin{abstract}
We present a Bayesian approach to adapting parameters of a well-trained context-dependent, deep-neural-network, hidden Markov model (CD-DNN-HMM) to improve automatic speech recognition performance. Given an abundance of DNN parameters but with only a limited amount of data, the effectiveness of the adapted DNN model can often be compromised. We formulate \emph{maximum a posteriori} (MAP) adaptation of parameters of a specially designed CD-DNN-HMM with an augmented linear hidden networks connected to the output tied states, or senones, and compare it to feature space MAP linear regression previously proposed. Experimental evidences on the 20,000-word open vocabulary Wall Street Journal task demonstrate the feasibility of the proposed framework. In supervised adaptation, the proposed MAP adaptation approach provides more than 10\% relative error reduction and consistently outperforms the conventional transformation based methods. Furthermore, we present an initial attempt to generate hierarchical priors to improve adaptation efficiency and effectiveness with limited adaptation data by exploiting similarities among senones.

 %Furthermore, it  outperforms already very strong speaker independent CD-DNN-HMM systems using different adaptation sets .

\end{abstract}

\section{Introduction}
\label{sec:intro}
Despite the recent outstanding results demonstrated by context-dependent, deep-neural-network based hidden Markov models (CD-DNN-HMMs) in various automatic speech recognition (ASR) tasks and data sets \cite{sainath2011making,dahl2012context,vesely2013sequencediscriminative}, these acoustic models, similarly to conventional context-dependent, Gaussian-mixture-model based HMMs (CD-GMM-HMMs) \cite{rabiner1989tutorial}, still suffer from a performance degradation under potential mismatched conditions between training and testing conditions. For standard hybrid system using  artificial neural networks (ANNs) and HMMs \cite{Bourlard94} in which CD-DNN-HMM is a special case, there exist many adaptation techniques. The simplest approach modifies all weights of the connectionist architecture using some adaptation materials. Unfortunately, it leads to over-fitting on the adaptation material when the amount of adaptation patterns is limited \cite{neto1995speaker}. Recent approaches, such as regularization based \cite{li2006regularized, yu2013kl}, subspace based \cite{yu2012factorized, yu2013deep},  transformation based \cite{neto1995speaker,gemello2007linear,seide2011feature,yao2012adaptation}, i-Vector based \cite{Saon2013}, native neural network based \cite{siniscalchi2013hermitian, Abdel-Hamid2013, Swietojanski2014}, factorization based \cite{Li14factorized} and fast adaptation schemes based on discriminant speaker codes \cite{Abdel-Hamid2013b, Abdel-Hamid2013, xue2014direct, Xue2014}, have been proposed to circumvent the problem. %The latter approach has proven to be very powerful in several speech tasks (see \cite{Xue2014}), yet it requires the pre-training of additional connection weights between speaker codes and hidden units.

Transformation based methods are the most popular connectionist adaptation techniques. The key idea is to augment the structure of the ANN component by adding an affine transformation network to the input \cite{neto1995speaker}, hidden \cite{gemello2007linear}, or output layer \cite{Sim2010}. They are typically trained while keeping the rest of the network parameters fixed. Motivations for these approaches stem from the concept that only relatively few parameters could be learned during adaptation and therefore it is preferable to training the entire network when the adaptation set is limited. For linear hidden network (LHN) layer approaches, the last hidden layer is usually designed to be a bottleneck to ensure an affordable parameter size \cite{sainath2013low,xue2013restructuring,xue2014singular,xue2014speaker}.

However, adapting parameters in a CD-DNN-HMM is much more challenging than earlier connectionist adaptation schemes because of its huge parameter set size with a large number of network branches connected to a large set of tied HMM states, often referred to as senones \cite{hwang1993shared}.
Furthermore, DNN parameters are adapted by every sample frame regardless of its senone class. Therefore, the posterior probabilities for the unobserved and scarcely seen senones are often pushed towards zero during adaptation. %Other senone-based states may only be observed a few times causing overfitting even when linear transformation approaches are used.
 Such a phenomenon is commonly referred to as \emph{catastrophic forgetting} \cite{Franch1994}.  Conservative ad-hoc solutions for ANNs have been proposed to force the senone distribution estimated from the adapted model to be close to that of the unadapted model.
For example, a Kullback-Leibler divergence (KLD) based objective criterion to be used during adaptation was devised in \cite{yu2013kl} in order to alleviate the catastrophic forgetting problem. A variation to the standard method of assigning the target values was instead discussed in \cite{gemello2007linear}. Nonetheless, Bayesian solutions adopted in the CD-GMM-HMMs to address the same issue \cite{Lee2000} have not been fully exploited.

In this study, we attempt to cast DNN adaptation within a Bayesian framework in the spirit of maximum a posteriori (MAP) adaptation \cite{gauvain1994maximum}. The key goal is to re-estimate some DNN parameters by representing available information in an augmented linear hidden network (LHN) added after the last non-linear hidden layer. Experimental results on the 20,000-word open vocabulary Wall Street Journal task demonstrates the feasibility of the proposed approach. Under supervised adaptation, the proposed MAP adaptation scheme can provide a relative word error rate (WER) reduction of more than 10\% from an already-strong speaker independent CD-DNN-HMM baseline and consistently outperform conventional transformation based adaptation schemes. It also compares favorably against the feature space maximum a posteriori linear regression approach to speaker adaptation proposed in \cite{huang2014feature}. We also present an initial attempt to generate hierarchical priors for improving adaptation efficiency with small amounts of adaptation data by exploiting the similarities among senones.

%The rest of the paper is organized as follows. A brief review of DNN training is given in Section \ref{sec:dnn}. Section \ref{sec:trans} discusses about the transformation based adaptation approaches. The proposed MAP adaptation technique is discussed in Section \ref{sec:map}. Experimental setup is presented in Section \ref{sec:results} along with experimental results. Section \ref{sec:conclusion} concludes our work.

\section{Training of Deep Models}
\label{sec:dnn}
%In the following sections, a brief review of DNN training is first presented. Next, DNN adaptation based on linear transformation networks is described.

%\subsection{DNN Training}
%\label{sec:training}
%In this work, the input to DNN is a splice of a central frame (whose label is that for the splice) and its $n$ context frames on both left and right sides, \textit{e.g.,} $n=10$.
%The basic DNN structure is shown in Fig.~\ref{fig:nn}.
In DNNs, hidden layers are usually constructed by sigmoid units, and the output layer is a softmax layer. The values of the nodes can therefore be expressed as:
  \begin{eqnarray}
  	\mathbf{x}_{i} &=&
  	\begin{cases}
  		\mathbf{W}_1  \mathbf{o}^t + \mathbf{b}_1, & i = 1   \\
  		\mathbf{W}_i \mathbf{y}_{i-1} + \mathbf{b}_i, & i > 1
  		\label{eq:linear}
  	\end{cases}, \\
  	\mathbf{y}_{i} &=&
  	\begin{cases}
  		\mathrm{sigmoid} (\mathbf{x}_i), & i < L \\
  		\mathrm{softmax} (\mathbf{x}_i), & i = L
  		\label{eq:nonlinear}
  	\end{cases},
  \end{eqnarray}

\noindent where $\mathbf{W}_1$, and $\mathbf{W}_i$ are the weight matrices, $\mathbf{b_1}$, and $\mathbf{b_i}$ are the bias vectors, $\mathbf{o}^t$ is the input frame at time $t$, $L$ is the total number of the hidden layers, and both  $\mathrm{sigmoid}$ and $\mathrm{softmax}$ functions are element-wise operations. The vector $\mathbf{x}_i$ corresponds to pre-nonlinearity activations, and $\mathbf{y}_i$ and $\mathbf{y}_L$ are the vectors of neuron outputs  at the $i^{\mathrm{th}}$ hidden layer and the output layer, respectively.
The softmax outputs were considered as an estimate of the senone posterior probability:
  \begin{equation}
  	\label{eq:softfunc}
  	p(C_j|\mathbf{o}^t)= \mathbf{y}^t_L(j) = \frac{\mathrm{exp}(\mathbf{x}^t_L(j)) }{\sum\limits_{i}{\mathrm{exp}(\mathbf{x}^t_L(i))}},
  \end{equation}
where $C_j$ represents the $j^{\mathrm{th}}$ senone and $\mathbf{y}_L(j)$ is the $j^{\mathrm{th}}$ element of $\mathbf{y}_L$.%in Fig.~\ref{fig:nn}.

%%\begin{figure}[t]
%%  \centering
%%  %\centerline{\includegraphics[width=0.8\linewidth]{nn}}
%%  \includegraphics[width=0.9\linewidth]{dnn-crop}
%%  \caption{Structure of a deep neural network: $W_i$ is the weight matrix at the $i^{\mathrm{th}}$ layer, omitting the bias terms for simplicity.}
%%  \label{fig:nn}
%%\end{figure}

The DNN is trained by maximizing the log posterior probability over the training frames. This is equivalent to minimizing the cross-entropy objective function. Let $\mathcal{X}$ be the whole training set, which contains $T$ frames, \textit{i.e.} $\mathbf{o}^{1:T} \in \mathcal{X}$, then the loss with respect to $\mathcal{X}$ is given by
  \begin{equation}
  	\mathcal{L}^{1:T} = - \sum_{t=1}^T \sum_{j=1}^J  \mathbf{\tilde{p}}^t (j) \log p(C_j|\mathbf{o}^t),
  	\label{eq:xent}
  \end{equation}
where $p(C_j|\mathbf{o}^t)$ is defined in Eq.~(\ref{eq:softfunc}); $\mathbf{\tilde{p}}^t$ is the target probability of frame $t$. In real practices of DNN systems, the target probability $\mathbf{\tilde{p}}^t$ is often obtained by a forced alignment with an existing system resulting in only the target entry that is equal to 1. Mini-batch  stochastic gradient descent (SGD) \cite{dekel2011optimal}, with a reasonable size of mini-batches to  make all matrices fit into the GPU memory, was used to update all neural parameters during training. Pre-training methods was used for the initialisation of the DNN parameters \cite{hinton2006reducing}.

%\subsection{KLD Adaptation for Affine Transformation}
%\label{sec:training}

\section{Transformation Based Adaptation for Deep Models}
\label{sec:trans}

%The hidden layers of a DNN can be viewed as extracting some abstract features from the input for classification. The activation values of the last hidden layer represent an internal structure of the input patterns in a space more suitable for classification and the output layer serves as a log-linear model taking the input of the last hidden layer.

For DNN adaptation, some researchers choose to add an affine transformation network between the last hidden layer and the output layer weights matrix, i.e., an LHN, and adapt only the LHN parameters while keeping fixed all of the other DNN parameters \cite{gemello2007linear}. In order to reduce the amount of parameters to adapt, usually the last hidden layer is designed to be a bottleneck (less neurons) \cite{sainath2013low,xue2013restructuring,xue2014singular,xue2014speaker}. Superior results were obtained by this kind of LHN formulation than other transformation based adaptation schemes, such as linear input network (LIN) and linear output network (LON).

The LIN approach performs adaptation by adding an augmented linear input layer and only adapts this set of LIN parameters. If we follow the common idea that the hidden layers of a DNN is actually learning a more suitable data representation and extracting better ``feature" for the output layer that is serving as a log-linear model, then by transforming the raw input using the LIN layer, we might harm the ability of data representation of the hidden layers. On the other hand for the LON approach, the issue is that usually we can't reduce the number of neurons of the output layer because we want to directly model the senones (the number of senones can be more than 10000 in practice), and that means we have to add a huge augmented layer with even more parameters to be adapted.

If we deem the hidden layers as a feature extractor and the output layer as the discriminative model. The model parameters are the weights of the output layer's affine transform matrix, $\mathbf{W}_L$. The output $\mathbf{y}_L$ can now be expressed as:

\begin{equation}
\mathbf{y}_L = \mathrm{softmax}(\mathbf{W}_{L}\mathbf{y}_{L-1}),
\label{eq:output}
\end{equation}
where the activation at the last hidden layer, $\mathbf{y}_{L-1}$, can be used as the new feature representation extracted by the hidden layers. When adding an augmented LHN after the last hidden layer, it is equivalent to applying a transformation matrix $\mathbf{W}_{lhn}$ to the model parameters to obtain an adapted model parameter set:

\begin{equation}
\mathbf{y}_L = \mathrm{softmax}(\mathbf{W}_{lhn}\mathbf{W}_{L}\mathbf{y}_{L-1}),
\label{eq:output_trans}
\end{equation}

\begin{figure}[t]
  \centering
  %\centerline{\includegraphics[width=0.8\linewidth]{nn}}
  \includegraphics[width=\linewidth]{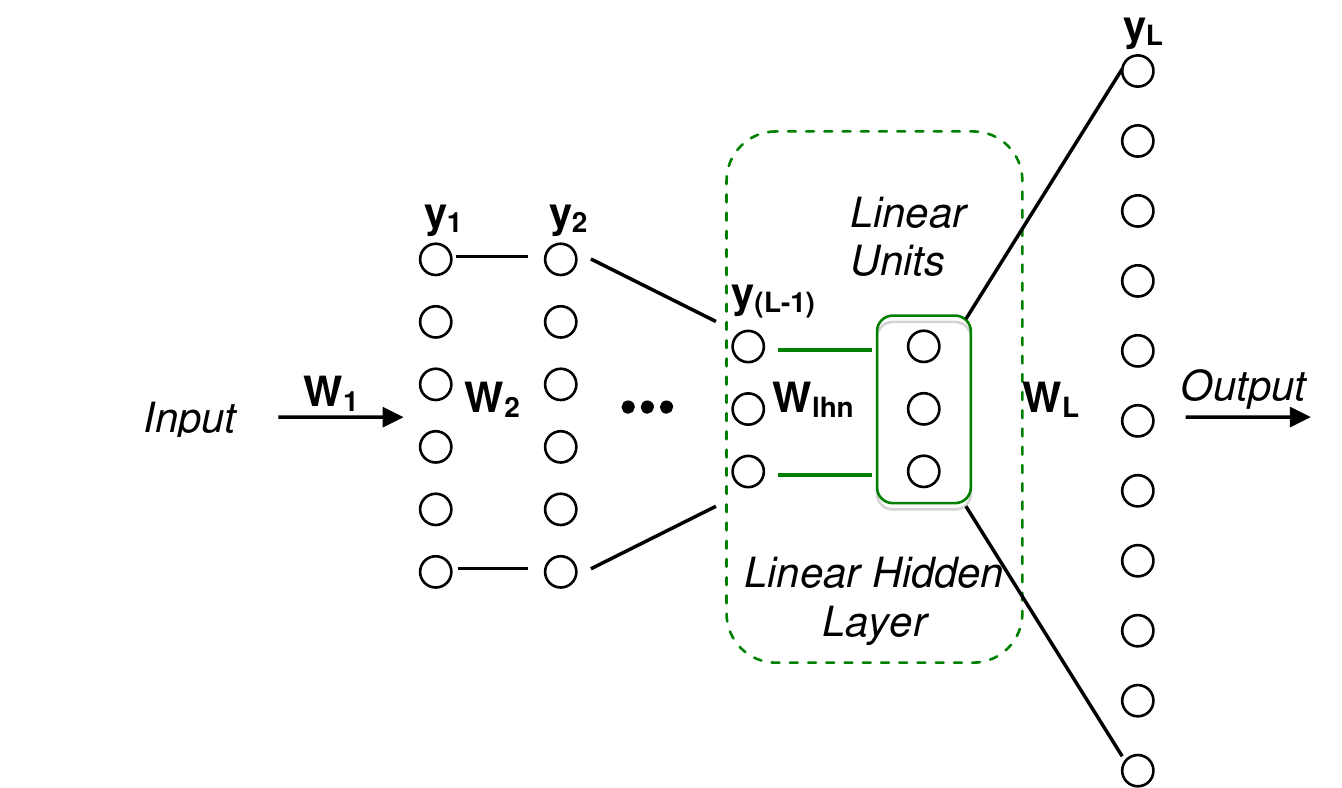}
  \caption{Basic neural architecture for adapting the HMM/ANN parameters: weights associated with the links in the dashed rectangles are estimated while all other weights remain unchanged. The activation function of each LHN units is a linear function.}
  \label{fig:lhn}
\end{figure}

An LHN adaptation structure is shown in Figure \ref{fig:lhn}. This formulation is quite similar to maximum likelihood linear regression (MLLR) \cite{leggetter1995maximum}. The difference is that in MLLR the model parameters are Gaussian mean and variance while here the model parameters are the log-linear model's transformation matrix weights.

%
% To contrast catastrophic forgetting during the adaptation phase, we follow \cite{yu2013kl} and use the KLD objective function  given in Eq. (\ref{eq:xentKLD}):
%
%\begin{equation}
%\mathcal{L}_{1:N} = - \sum_{t=1}^N \sum_{j=1}^J  {\hat{p}}_t (j) \log p(C_j|\mathbf{o}_t),
%\label{eq:xentKLD}
%\end{equation}
%where $\hat{p}_t$ is equal to $(1-\rho)(\tilde{p}(C_j | \mathbf{o}_t)+\rho p^{DNN}(C_j | \mathbf{o}_t)$, and $p^{DNN}(C_j | \mathbf{o}_t)$ is he posterior probability estimated from the unadapted DNN. Therefore, the target probability distribution has been changed to a linear interpolation of the distribution estimated from the unadapted DNN and the ground throughout alignment of the adaptation data (see \cite{yu2013kl} for more details). In this work, the affine transformation network is  estimated using $\rho$ set to 0.5.

%\begin{figure}[t]
%  \centering
%  \includegraphics[width=8cm]{weights}
%  \caption{histograms of 4 sample weights}
%  \label{fig:weights}
%\end{figure}

\section{MAP Adaptation for Deep Models}
\label{sec:map}

Although  conventional DNN adaptation approaches try to  alleviate over-fitting issues by reducing the number of parameters to be adapted, such number could still be very big in some cases. Inspired by the MAP adaptation that address the problem effectively in GMM-HMM systems, in this section, we explain how to apply the MAP approach to the LHN adaptation. Note that though we choose LHN for demonstration, the proposed MAP approach can be easily applied to other DNN adaptation frameworks like \cite{yu2013kl, Saon2013, Abdel-Hamid2013, Swietojanski2014} as well.

\subsection{Prior Estimation}
\label{ssec:prior_est}
In order to establish a MAP adaptation framework like in \cite{gauvain1994maximum}, a prior distribution over the weights of the affine transformation network need to be imposed. To analyze and estimate the prior density, we utilized the training data of the baseline DNN. We adopted an empirical Bayes approach \cite{gauvain1994maximum, Lee2000} and treated each speaker in the training set as a sample speaker and supervised LHN adaptation was performed. After that, we can get a particular LHN for each speaker. We observed that the histograms for weights of the adapted LHN over speakers are quite like Gaussian, so we assume that the distribution of the weights in $\mathbf{W}_{lhn}$ to be joint Gaussian \cite{huang2014feature}. By expressing the weights in the LHN transformation matrix $\mathbf{W}_{lhn}$ as a vector $\mathbf{w}$ with each entry representing a particular weight, we have the prior density in the following form:
\begin{equation}
p(\mathbf{W}_{lhn}) = \frac{1} {(2\pi)^{M/2}|\mathbf{\Sigma}|^{1/2}}\exp(- \frac{1}{2}(\mathbf{w}-\boldsymbol{\mu})^T\mathbf{\Sigma}^{-1}(\mathbf{w}-\boldsymbol{\mu}))
\label{eq:prior}
\end{equation}
where only the diagonal entries of the covariance matrix $\Sigma$ are non-zero (from the independence assumption of the weights).

With $N$ adapted speaker weight vectors, the maximum likelihood estimation of the mean $\boldsymbol{\mu}$ and variance $\boldsymbol{\Sigma}$ can be expressed as:

\begin{equation}
\boldsymbol{\mu}_{ML} =  \frac{1}{N}\sum_{i=1}^N \mathbf{w}_i
\label{eq:gmean}
\end{equation}

\begin{equation}
\mathbf{\Sigma}_{ML} =  \frac{1}{N}\sum_{i=1}^N (\mathbf{w}_i-\boldsymbol{\mu}_{ML})(\mathbf{w}_i-\boldsymbol{\mu}_{ML})^T
\label{eq:gvar}
\end{equation}
where $\mathbf{w}_i$ is the vector consisting of the adapted transformation weights of speaker $i$.

\subsection{MAP Formulation}
\label{ssec:formal_map}

Formal MAP adaptation is conducted following  \cite{huang2014feature}. Eq.~(\ref{eq:xent_scalor}) formulates the MAP learning idea by adding the term of prior density $p(\mathbf{W}_{lhn})$ to the plain cross entropy objective function.

\begin{equation}
\mathcal{L}^{1:T}_{MAP}  = -\lambda\log p(\mathbf{W}_{lhn})+\mathcal{L}^{1:T}_{xent}
\label{eq:xent_scalor}
\end{equation}
Applying the prior form of Eq. (\ref{eq:prior}), the objective function for MAP LHN adaptation is in the form of Eq. (\ref{eq:xent_w}).

 \begin{equation}
\mathcal{L}^{1:T}_{MAP} =\frac{ \lambda}{2}(\mathbf{w}-\boldsymbol{\mu})^T\boldsymbol{\Sigma}^{-1}(\mathbf{w}-\boldsymbol{\mu}) +\mathcal{L}^{1:T}_{xent}
\label{eq:xent_w}
\end{equation}
where only the diagonal entries of the covariance matrix $\boldsymbol{\Sigma}$ are non-zero (from the independence assumption of the weights).

A close look at Eq.~(\ref{eq:xent_w}), when the prior density is a standard Gaussian $\mathcal{N}(\mathbf{0},\mathbf{I})$, MAP learning will degenerate to conventional L2-regularized training. The gradient of $\mathcal{L}_{1:N}^{MAP}$ with respect to $\mathbf{w}$ can now be expressed as:
\begin{equation}
\frac{\partial \mathcal{L}^{1:T}_{MAP}}{\partial \mathbf{w}} =\lambda (\mathbf{w}-\boldsymbol{\mu})^T diag(\boldsymbol{\Sigma}^{-1}) + \frac{\partial \mathcal{L}^{1:T}_{xent}}{\partial \mathbf{w}},
\label{bdescent2}
\end{equation}
where $diag(\mathbf{\Sigma}^{-1})$ consists of the diagonal entries of $\mathbf{\Sigma}^{-1}$.

\section{Experiments}
\label{sec:results}
\subsection{Experimental Setup}
\label{sec:setup}
This study is concerned with the problem of \emph{speaker adaptation}, and experiments are reported on the 20k-word open vocabulary Wall Street Journal task \cite{Paul1992} using the Kaldi toolkit \cite{povey2011kaldi}.  The baseline CD-DNN-HMM system was trained using the WSJ0 material (SI-84). The standard adaptation set of WSJ0 (si\_et\_ad, 8 speakers, 40 sentences per speaker) was used to perform adaptation of the affine transformation added to the speaker-independent DNN. The standard open vocabulary 20,000-word (20K) read NVP Senneheiser microphone (si\_et\_20, 8 speakers x ~40 sentences) data were used for evaluation. A standard trigram language model was adopted during decoding. The ASR performance was given in terms of the word error rate (WER).

The DNN has six hidden layers. The first five hidden layers have 2048 units, whereas the last hidden layer has 216 units. The output layer has 2022 softmax units. %corresponding to the senones generated using a CD-GMM-HMM baseline.
This DNN architecture follows  conventional configurations used in the speech community except for the last hidden layer, which acts as a bottleneck layer. This configuration was chosen, because a too large  dimension of the last non-linear hidden layer might have been harmful for LHN adaptation. The bottleneck based low rank methods have been widely used to achieve more compact DNN models with equivalent performance \cite{sainath2013low,xue2013restructuring,xue2014singular,xue2014speaker}. The number units equal to 216 was chosen to simulate a sort of three-state phone layer thereby obtaining a kind of hierarchical structure between mono-phones in the hidden layer and senones at the output layer. The input feature vector is a 23-dimension mean-normalized log-filter bank feature with up to second-order derivatives and a context window of 11 frames, forming a vector of 759-dimension ($69\times11$) input.  The DNN was trained with an initial learning rate of 0.008 using the cross-entropy objective function. It was initialised with the stacked restricted Boltzmann machines  by using layer by layer generative pre-training.% An initial learning rate of 0.01 was then used to train the Gaussian-Bernoulli RBM and a learning rate of 0.4 was applied to the Bernoulli-Bernoulli RBMs.

\subsection{Experimental Results}
\label{sec:experiments}
The word error rate (WER) attained with different adaptation techniques are reported  in Table \ref{exp:comparison}. All available adaptation material was used for performing adaptation, namely 40 sentences per speaker. The term BASELINE refers to the speaker independent CD-DNN-HMM system. LIN, LIN-KLD, and MAP LIN refer to the adaptation technique based on the standard linear input network approach,  the  KLD regularisation technique\footnote{The best KLD results obtained in our laboratories are reported.} in combination with LIN, and the maximum a posteriori transformation based adaptation when  a prior is defined over the LIN parameters \cite{huang2014feature}, respectively. The terms LON and LON-KLD are used to denote,   with a little abuse of terminology, the direct adaptation of the output layer weights matrix with or without  KLD, respectively. LHN adaptation results  are also reported along with the corresponding MAP version, MAP LHN, which is  the adaptation approach proposed in this paper. Since LHN was inserted between the last hidden layer  and the output layer weights matrix, its dimension is $216\times216$. LHN is initialised to an identity matrix with zero bias, which gives a starting point  equivalent to the unadapted model.  Supervised adaptation is then performed updating only the LHN parameters. % The learning rate was set to 0.0002 in all experiments.
MAP was performed as described in Section \ref{sec:map}.  For the sake of comparison,  LHN-KLD, which denotes  standard  LHN  combined with KLD, was also evaluated.

Indeed LIN and LHN outperforms LON, which attains the worst performance improvement.  KLD always improves over affine transformation based adaptation techniques, as expected. Furthermore, the proposed MAP LHN outperforms all other techniques in the given task, and it attains the best recognition results with a relative improvement of 10.4\% over the BASELINE. Finally, we  would like to remark that  MAP LHN compares favourably against MAP LIN, and that confirms that  the introduction of the bottleneck layer was the key for a proper deployment of MAP LHN.

 \begin{table}[t]\footnotesize
\begin{center}
  \caption{{Comparing WERs on the 20K word open vocabulary WSJ0 task for several adaptation approaches using all 40 available adaptation utterances.}}
  \label{exp:comparison}
  \centerline{
  \begin{tabular}{|c||c|}
  \hline
%\pbox{10cm}{{\bf Affine Transformation}\\ (\#parameters)} & {\bf System} & {\bf WER (\%)} & Rel. Impr. \\
{\bf System} & {\bf WER (in \%)}  \\
\hline
\hline
BASELINE &  \multicolumn{1}{c|}{8.84\%}\\
  \hline
  \hline
{\em LIN} &  8.22\%  \\
 \hline
{\em LIN-KLD}  &  8.06\% \\
 \hline
{\em MAP LIN}  &   8.13\%\\
 \hline
 \hline
{\em LON} &   8.80\%  \\
 \hline
{\em LON-KLD}  &  8.64\% \\
 \hline
 \hline
{\em LHN} &   8.22\%  \\
 \hline
{\em LHN-KLD}  &  8.15\% \\
 \hline
{\em MAP LHN}  &   {\bf 7.92}\%\\
 \hline
 \hline
  \end{tabular}}
  \vspace{-6mm}
 \end{center}
  \end{table}

Table \ref{exp:adpt} shows experimental results for LHN, and MAP LHN with  different amounts of adaptation sentences, namely $\{5, 10, 20, 40\}$, in the second and third columns, respectively. These results confirm that MAP LHN adaptation almost always improves over standard LHN, with the best adaptation results at a WER of 7.92\% using 40 utterances. But in very limited adaptation data cases, namely, $\{5, 10\}$, there is only slight or even no improvement by only using flat prior in the MAP adaptation, so we turned to our preliminary investigation of hierarchical priors for dealing with the data scarcity problem.

 \begin{table}[t]\footnotesize
\begin{center}
  \caption{{Comparing LHN and MAP LHN performance with different amounts of adaptation data.}}
  \label{exp:adpt}
  \centerline{
  \begin{tabular}{|c||c|c|}
  \hline
%\pbox{10cm}{{\bf Affine Transformation}\\ (\#parameters)} & {\bf System} & {\bf WER (\%)} & Rel. Impr. \\
\vtop{\hbox{\strut \# Adaptation}\hbox{\strut \,\,\,\,\,Sentences}} & {\vtop{\hbox{\strut Standard}\hbox{\strut \,\,\,\,\,LHN}}} & {\vtop{\hbox {\strut  MAP}\hbox{\strut  LHN}}} \\
\hline
BASELINE &  \multicolumn{2}{c|}{8.84\%}\\
\hline
{\em 5}  &  8.59\% & 8.54\% \\
\hline
{\em 10} &  8.52\% & 8.52\% \\
\hline
{\em 20}  &  8.31\% & 8.12\% \\
\hline
{\em 40} &  8.22\% & 7.92\% \\
 \hline
  \end{tabular}}
  \vspace{-6mm}
 \end{center}
  \end{table}
  
\subsection{Hierarchical Priors: Preliminary Experiments}
\label{ssec:hier_prior}

\begin{figure}[t]
	\centering
	%\centerline{\includegraphics[width=0.8\linewidth]{nn}}
	\includegraphics[width=0.8\linewidth]{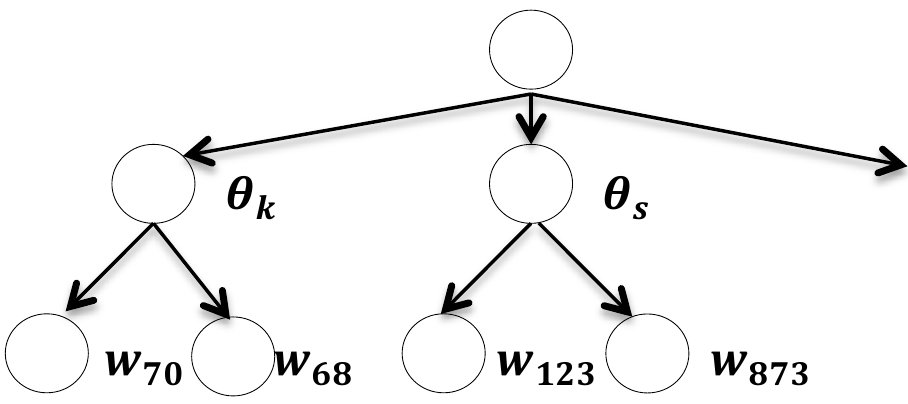}
	\caption{Fixed two-layer tree for hierarchical priors generation. Each leaf node represents a senone embedding, and it is thereby a row of  $\mathbf{W}_{(D +1) \times L}$.}
	\label{fig:tree}
	 \vspace{-4mm}
\end{figure}

Hierarchical structures, such as trees, have long been used in the speech community to address the over-fitting issues during model parameters estimation. For instance,  efficient adaptation with a limited amount of adaptation data was obtained through the use of  a tree data structure to cluster model parameters of a CD-GMM-HMM system in \cite{shinoda2001structural}. Similar ideas have been recently explored in  DNN learning for enhancing classification performance for classes with few examples in \cite{Srivastava2013},  where hierarchical priors where devised for the output layer weights matrix (top-level weights in a DNN) using a tree data structure either fixed or learnable during training.
% It may be useful at this point to recall that the process to obtain posterior probabilities of senone classes in CD-DNN-HMMs can be seen as a two-stage procedure: in the first stage, the first layers map the input observation vector into a high-level feature vector; the posterior probability are then obtained using a log-linear model with the high-level feature from the last hidden layer in the second stage.

Top-level DNN weights in a hybrid acoustic model can  be regarded as  senone embeddings \cite{XLi2014},  and hierarchical priors can be defined by organising those embedding in a tree data structure. Let $\mathbf{W}_{(D+1) \times L}$ denote the output layer weights matrix (including the bias terms). Each line in $\mathbf{W}_{(D+1) \times L}$  corresponds to a senone embedding. Specifically, the $s$th senone embedding can be denoted as $\mathbf{w}_{s}$, which is the $s$th row in $\mathbf{W}_{(D+1) \times L}$.  The tree structure used to generate hierarchical priors can be either learnt during training or given.  Here, we used a fixed two-layer  tree shown in Figure \ref{fig:tree}: there are $L$ leaf nodes, with each leaf corresponding to a senone embedding, and $S$ parent nodes clustering together similar leaf nodes. Each parent node clusters senone embeddings sharing the same central phone-state; therefore,  $S$ is equal to 130 in this work. Hierarchical priors can now be established by associating a vector $\mathbf{w}_{s}$ to a leaf node, and a vector $\boldsymbol{\theta}_{s}$ to each parent node, and imposing a  Gaussian probability density distribution over these two vectors as follows: $\boldsymbol{\theta}_{s} \sim \mathcal{N}(\mathbf{0},\frac{1}{\sigma_{1}^{2}} \mathbf{I}_{(D+1)})$, and $\mathbf{w}_{s} \sim  \mathcal{N}(\boldsymbol{\theta}_{s},\frac{1}{\sigma_{2}^{2}} \mathbf{I}_{(D+1)})$.

The objective function with hierarchical priors is in the form of Eq. (\ref{eq:xent_hier}).

\begin{equation}
\mathcal{L}_{1:N}^{MAP} =\mathcal{L}_{1:N}^{xent} + \frac{\lambda_{2}}{2} \sum \lVert  \mathbf{w}_{s} - \boldsymbol{\theta}_{s} \rVert^{2} +  \frac{\lambda_{1}}{2} \lVert \boldsymbol{\theta}  \rVert^{2}.
\label{eq:xent_hier}
\end{equation}

 \begin{table}[t]\footnotesize
 	\begin{center}
 		\caption{{WER comparisons of flat and hierarchical priors for MAP LHN with a small amount of adaptation utterances.}}
 		\label{exp:hier}
 		\centerline{
 			\begin{tabular}{|c||c|c|}
 				\hline
 				%\pbox{10cm}{{\bf Affine Transformation}\\ (\#parameters)} & {\bf System} & {\bf WER (\%)} & Rel. Impr. \\
 				\vtop{\hbox{\strut \# Adaptation}\hbox{\strut \,\,\,\,\,Sentences}} & {\vtop{\hbox {\strut  MAP LHN}\hbox{\strut  Flat Priors}}} & {\vtop{\hbox{\strut \,\,\,\,\,\,\, MAP LHN}\hbox{\strut Hierarchical Priors}}} \\
 				\hline
 				{\em 5} &   8.54\%  & 8.48\% \\
 				\hline
 				{\em 10}  &  8.52\% & 8.45\% \\
 				\hline
 			\end{tabular}}
 			\vspace{-5mm}
 		\end{center}
 	\end{table}

It is can be verified that $\boldsymbol{\theta}_{s}$ is a scaled average of all $\mathbf{w}_{s}$ associated to the $s$th leaf node by minimizing Eq. \ref{eq:xent_hier} over $\boldsymbol{\theta}_{s}$  with fixed DNN weights (see \cite{Srivastava2013}). We focus our attention on experimental results with very small adaptation data amounts, as shown in Table \ref{exp:hier}. With limited adaptation data, namely {5, 10} utterances, small performance improvements are observed against using flat priors when adaptation is carried out with hierarchical priors. Although the current improvement is still quite small, we believe more sophisticated trees can be adopted for better performance in future studies. %This goal is reached by combining  $W_{lhn}$  and  $W_{n}$, see Figure \ref{fig:lhn}, into a single matrix $W$ through simple matrix manipulations: $W = W_{n} \times W_{lhn}$, and $bias = bias_{n} + W_{n} \times bias_{n}$.

\section{Conclusion}
\label{sec:conclusion}
We have investigated a \emph{maximum a posteriori} (MAP) adaptation approach for linear hidden networks. The key idea is to treat the parameters of the augmented affine transformation as random Gaussian variables and incorporate prior information obtained from the training data. Speaker adaptation results show that the proposed MAP approaches can lead to a consistent performance improvement over conventional LHN adaptation. Furthermore, MAP LHN outperforms other regularisation schemes. 

A first attempt to use hierarchical-based priors with a fixed two-layer tree structure was also studied, and small improvements were observed in a set of preliminary ASR experiments using a limited amount of adaptation sentences. Better results might still be hindered by the current fixed tree hierarchy  structure employed in this preliminary work. Indeed, it was demonstrated that learning the tree hierarchy during training improves the classification performance \cite{Srivastava2013}.  Finally, from the objective function perspective, we are still relying on cross-entropy. Other forms of frame-level and sequence-level discriminative objectives \cite{vesely2013sequencediscriminative,huang2014beyond} can also be applied.

%Finally, another important issue that should be addressed is that all parameters are updated for each adaptation input frame, which
%\section{REFERENCES}
%\label{sec:refs}


 \clearpage

  \eightpt
  \bibliographystyle{IEEEtran}
\bibliography{refs}

\end{document}